%% file: main.tex
\documentclass[11pt,a4paper]{article}
\usepackage[hyperref]{acl2019}
\usepackage{times}
\usepackage{latexsym}
\usepackage{amsmath}
\usepackage{url}
\usepackage{graphicx}
\usepackage{subfigure}
\usepackage{siunitx}
\usepackage{booktabs}
\usepackage{etoolbox}

\usepackage{todonotes}
\usepackage[export]{adjustbox}

\aclfinalcopy 
\def\aclpaperid{P19-1506} 

\title{Like a Baby: Visually Situated Neural Language Acquisition}

\author{Alexander G. Ororbia*$^{1,2}$, Ankur Mali *$^{1,2}$, Mary Alexandria Kelly$^1$, \and David Reitter$^{1,3}$ \\
  (1) The Pennsylvania State University, University Park, PA, USA \\
  (2) Rochester Institute of Technology, Rochester, NY, USA \\
  (3) Google Research, New York City, NY, USA \\
  {\tt  ago@cs.rit.edu, aam35@psu.edu,} \\
  {\tt mary.kelly4@carleton.ca, reitter@google.com}}
  
\date{}

\begin{document}
\maketitle

\pagestyle{plain}
\pagenumbering{arabic}
\setcounter{page}{5127}

\AddToShipoutPicture*{
  \setlength{\unitlength}{1mm}
  \footnotesize

  \put(0,13){\parbox[t]{\paperwidth}{\centering
        \emph{Proceedings of the 57th Annual Meeting of the Association for Computational Linguistics,}, pages 5127--5136 \\
        Florence, Italy, July 28 - August 2, 2019. \textcopyright
        2019 Association for Computational Linguistics}}
}

\begin{abstract}
We examine the benefits of visual context in training neural language models  to perform next-word prediction. A multi-modal neural  architecture is introduced that outperform its equivalent trained on language alone with a 2\% decrease in perplexity, even when no visual context is available at test. Fine-tuning the embeddings of a pre-trained state-of-the-art bidirectional language model (BERT) in the language modeling framework yields a 3.5\% improvement. The advantage for training with visual context when testing without is robust across different languages (English, German and Spanish) and different models (GRU, LSTM, $\Delta$-RNN, as well as those that use BERT embeddings). Thus, language models perform better when they learn like a baby, i.e, in a multi-modal environment. This finding is compatible with the theory of situated cognition: language is inseparable from its physical context.
\end{abstract}

\section{Introduction}

\label{sec:intro}
The theory of situated cognition postulates that a person's knowledge is inseparable from the physical or social context in which it is learned and used \citep{greeno1993situativity}. Similarly, Perceptual Symbol Systems theory holds that all of cognition, thought, language, reasoning, and memory, is grounded in perceptual features \cite{barsalou1999perceptions}. Knowledge of language cannot be separated from its physical context, which allows words and sentences to be learned by grounding them in reference to objects or natural concepts on hand (see \citeauthor{roy2005connecting}, \citeyear{roy2005connecting}, for a review). Nor can knowledge of language be separated from its social context, where language is learned interactively through communicating with others to facilitate problem-solving. Simply put, language does not occur in a vacuum. 

Yet, statistical language models, typically connectionist systems, are often trained in such a vacuum. Sequences of symbols, such as sentences or phrases composed of words in any language, such as English or German, are often fed into the model independently of any real-world context they might describe. In the classical language modeling framework, a model learns to predict a word based on a history of words it has seen so far. While these models learn a great deal of linguistic structure from these symbol sequences alone, acquiring the essence of basic syntax, it is highly unlikely that this approach can create models that acquire much in terms of semantics or pragmatics, which are integral to the human experience of language. How might one build neural language models that  ``understand'' the semantic content held within the symbol sequences, of any language, presented to it?

In this paper, we take a small step towards a model that understands language as a human does by training a neural model jointly on corresponding linguistic and visual data.
From an image-captioning dataset, we create a multi-lingual corpus where sentences are mapped to the real-world images they describe. We ask how adding such real-world context at training can improve language model performance. We create a unified multi-modal connectionist architecture that incorporates visual context and uses either $\Delta$-RNN \cite{ororbia2017learning}, Long Short Term Memory \cite[LSTM;][]{Hochreiter:1997} or Gated Recurrent Unit \cite[GRU;][]{cho2014learning} units. We find that the models acquire more knowledge of language than if they were trained without corresponding, real-world visual context. 

\section{Related Work}
\label{sec:related_work}
Both behavioral and neuroimaging studies have found considerable evidence for the contribution of perceptual information to linguistic tasks \cite{barsalou2008grounded}. It has long been held that language is acquired jointly with perception through interaction with the environment \citep[e.g.][]{frank2008bayesian}. Eye-tracking studies show that visual context influences word recognition and syntactic parsing from even the earliest moments of comprehension \cite{Tanenhaus1995}.

Computational cognitive models can account for bootstrapped learning of word meaning and syntax when language is paired with perceptual experience \citep{abend2017bootstrapping} and for the ability of children to rapidly acquire new words by inferring the referent from their physical environment \cite{alishahi2008fast}. Some distributional semantics models integrate word co-occurrence data with perceptual data, either to achieve a better model of language as it exists in the minds of humans \cite{Baroni2016visual,johns2012perceptual,kievit2011semantic,lazaridou2014wampimuk} or to improve performance on machine learning tasks such as object recognition \cite{frome2013devise,Lazaridou2015adjectives}, image captioning \cite{kiros2014unifying,lazaridou2015multimodal}, or image search \cite{socher2014grounded}. 

Integrating language and perception can facilitate language acquisition by allowing models to infer how a new word is used from the perceptual features of its referent \cite{johns2012perceptual} or to allow for \textit{fast mapping} between a new word and a new object in the environment \cite{lazaridou2014wampimuk}. Likewise, this integration allows models to infer the perceptual features of an unobserved referent from how a word is used in language \cite{johns2012perceptual,lazaridou2015multimodal}. As a result, language data can be used to improve object recognition by providing information about unobserved or infrequently observed objects \cite{frome2013devise} or for differentiating objects that often co-occur in photos \cite[e.g., cats and sofas;][]{Lazaridou2015adjectives}.

By representing the referents of concrete nouns as arrangements of elementary visual features \cite{biederman1987recognition}, \citet{kievit2011semantic} found that the visual features of nouns capture semantic typicality effects, and that a combined representation, consisting of both visual features and word co-occurrence data, more strongly correlates with human judgments of semantic similarity than representations extracted from a corpus alone. While modeling similarity judgments is distinct from the problem of predictive language modeling, we take this finding as evidence that visual perception informs semantics, which suggests there are gains to be had integrating perception with predictive language models.

In contrast to prior work in machine learning, where mappings between vision and language have been examined \cite{kiros2014unifying,vinyals2015show,xu2015show}, our goal in integrating visual and linguistic data is not to accomplish a task such as image search/captioning that inherently requires a mapping between these modalities. Rather, our goal is to show that, since perceptual information is intrinsic to how humans process language, a language model that is trained on both visual and linguistic data will be a better model, consistently across languages, than one trained on linguistic data alone.

Due to the ability of language models to constrain predictions on the basis of preceding context, language models play a central role in natural-language and speech processing applications. However, the psycholinguistic questions surrounding how people acquire and use linguistic knowledge are fundamentally different from the aims of machine learning. Using NLP language models to address psycholinguistic questions is a new approach that integrates well with the theory of predictive coding in cognitive psychology \citep{clark2013whatever,rao1999predictive}. For language processing this means that when reading text or comprehending speech, humans constantly anticipate what will be said next. Predictive coding in humans is a fast, implicit cognitive process similar to the kind of sequence learning that recurrent neural models excel at. We do not propose recurrent neural models as direct accounts of human language processing. Instead, our intent is to use a general purpose machine learning algorithm as a tool to investigate the informational characteristics of the language learning task. More specifically, we use machine learning to explore the question as to whether natural languages are most easily learned when situated in an environmental context and grounded in perception.


\section{The Multi-modal Neural Architecture }
\label{sec:arch}
We will evaluate the multi-modal training approach on several well-known complex architectures, including the LSTM, and further examine the effect of using pre-trained BERT embeddings. However, to simply describe the the neural model, we start from the Differential State Framework \citep[DSF;][]{ororbia2017learning}, which unifies gated recurrent architectures under the general view that state memory is a simple parametrized mixture of ``fast'' and ``slow'' states.
Our aim is to model sequences of symbols, such as the words that compose sentences, where at each time we process $\mathbf{x}_t$, or the one-hot encoding of a token\footnote{One-hot encoding represents tokens as binary-valued vectors with one dimension for each type of token. Only one dimension has a non-zero value, indicating the presence of a token of that type.}

\begin{figure*}[ht]
\begin{center}
\includegraphics[width=0.7\textwidth]{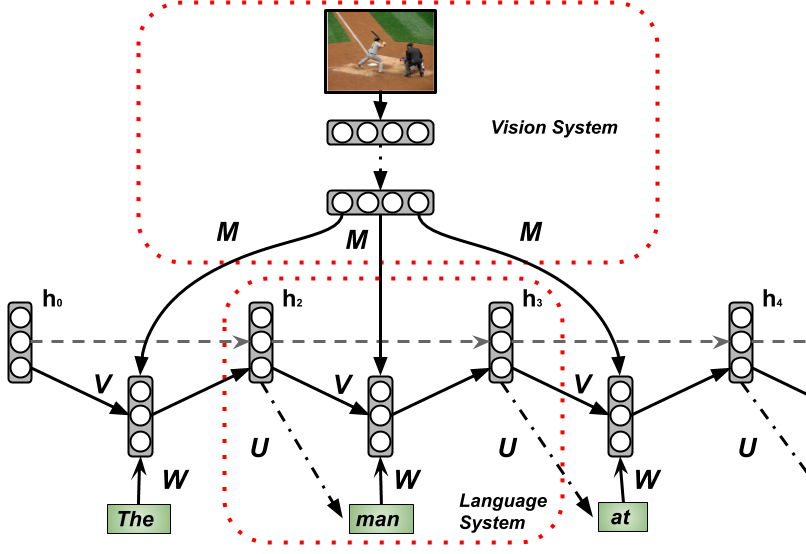}
\caption{Integration of visual information in an unrolled network (here, the MM-$\Delta$-RNN. Grey-dashed: identity connections; black-dash-dotted: next-step predictions; solid-back lines: weight matrices.}
\label{sysarch}
\end{center}
\end{figure*}

One of the simplest models that can be derived from the DSF is the $\Delta$-RNN \citep{ororbia2017learning}. A $\Delta$-RNN is a simple gated RNN that captures longer-term dependencies in sequences through the use of a parametrized, flexible state ``mixing'' function. The model computes a new state at a given time step by comparing a fast state (which is proposed after accounting for the current token) and a slow state (a form of long-term memory).
The model is defined by parameters $\Theta = \{W, V, \mathbf{b}_r, \beta_1, \beta_2, \alpha \}$ (input-to-hidden weights $W$, recurrent weights $V$, gating-control coefficients $\beta_1, \beta_2, \alpha$, and the rate-gate bias $\mathbf{b}_r$). Inference is defined as:
\begin{align}
\mathbf{d}^{rec}_t &= V \mathbf{h}_{t-1},\ \mathbf{d}^{dat}_t = W \mathbf{e}_{w,t}\\
\mathbf{d}^1_t &=  \alpha \otimes \mathbf{d}^{rec}_t \otimes \ \mathbf{d}^{dat}_t \\ \mathbf{d}^2_t &= \beta_1 \otimes \mathbf{d}^{rec}_t + \beta_2 \otimes \mathbf{d}^{dat}_t\\
\mathbf{z}_{t} &= \phi_{hid}(\mathbf{d}^1_t + \mathbf{d}^2_t) \\
\mathbf{h}_{t} &=  \Phi( (1 - \mathbf{r}) \otimes \mathbf{z}_{t} + \mathbf{r} \otimes \mathbf{h}_{t-1} )\\ \mathbf{r} &=  1 / (1 + exp(-[\mathbf{d}^{dat}_t + \mathbf{b}_r]))
\end{align}
where $\mathbf{e}_{w,t}$ is the 1-of-k encoding of the word $w$ at time $t$. Note that $\{\alpha, \beta_1, \beta_2\}$ are learnable bias vectors that modulate internal multiplicative interactions. The rate gate $\mathbf{r}$  controls how slow and fast-moving memory states are mixed inside the model.
In contrast to the model originally trained in \newcite{ororbia2017learning}, the outer activation is the linear rectifier, $\Phi(v) = max(0,v)$, instead of the identity or hyperbolic tangent, because we found that it worked much better. The inner activation function $\phi_{hid}(v)$ is $tanh(v) = \frac{(e^{(2v)} - 1)}{(e^{(2v)} + 1)}$.

To integrate visual context information into the $\Delta$-RNN, we fuse the model with a neural vision system, motivated by work done in automated image captioning \citep{xu2015show}. We adopt a transfer learning approach and incorporate a state-of-the-art convolutional neural network into the $\Delta$-RNN model, namely the Inception-v3 network \citep{szegedy2016rethinking}\footnote{In preliminary experiments, we also examined VGGNet and a few others, but found that the Inception worked the best when it came to acquiring more general distributed representations of natural images.}, in order to create a multi-modal $\Delta$-RNN model (MM-$\Delta$-RNN; see Figure \ref{sysarch}). 
Since our focus is on language modeling, the parameters of the vision network are fixed.


To obtain a distributed representation of an image from the Inception-v3 network, we extract the vector produced from the final max-pooling layer, $\mathbf{c}$, after running an image through the model (note that this operation occurs right before the final, fully-connected processing layers which are usually task-specific parameters, such as in object classification). The $\Delta$-RNN can make use of the information in this visual context vector if we modify its state computation in one of two ways. The first way would be to modify the inner state to be a linear combination of the data-dependent pre-activation, the filtration, and a learned linear mapping of $\mathbf{c}$ as follows: 
\begin{align}
\mathbf{z}_{t} &= \phi_{hid}(\mathbf{d}^1_t + \mathbf{d}^2_t + M \mathbf{c} + \mathbf{b})\label{eqn:inner_integration}
\end{align}
where $M$ is a learnable synaptic connections matrix that connects the visual context representation with the inner state. The second way to modify the $\Delta$-RNN would be change its outer mixing function instead:
\begin{align}
\mathbf{h}_{t} &=  \Phi( [ (1 - \mathbf{r}) \otimes \mathbf{z}_{t} + \mathbf{r} \otimes \mathbf{h}_{t-1} ] \otimes (M \mathbf{c}) ) \label{eqn:outer_integration}
\end{align}
Here in Equation \ref{eqn:outer_integration} we see the linearly-mapped visual context embedding interacts with the currently computation state through a multiplicative operation, allowing the visual-context to persist and work in a longer-term capacity.
In either situation, using a parameter matrix $M$ frees us from having to set the dimensionality of the hidden state to be the same as the context vector produced by the Inception-v3 network. 


We do not use regularization techniques with this model. The application of regularization techniques is, in principle, possible (and typically improves performance of the $\Delta$-RNN), but it is damaging to performance in this particular case, where an already compressed and regularized representation of the images from Inception-v3 serves as input to the multi-modal language modeling network.

Let $w_1, \dots, w_N$ be a variable-length sequence of $N$ words corresponding to an image $I$. In general, the distribution over the variables follows the graphical model:
\begin{align*}
P_{\theta}(w_1, \dots, w_T | I) = \prod_{t=1}^T P_{\Theta}(w_t | w_{<t}, I)
\end{align*}
For all model variants the state $\mathbf{h}_t$ calculated at any time step is fed into a maximum-entropy classifier\footnote{Bias term omitted for clarity.} defined as:
\begin{align*}
P(w,\mathbf{h}_t) = P_{\Theta}(w|\mathbf{h}_t) =  \frac{\exp{( w^{\text{T}} U \mathbf{h}_t)}}{\sum_{w'} \exp{((w')^{\text{T}} U \mathbf{h}_t)}} 
\end{align*}
The model parameters $\Theta$ optimized with respect to the sequence negative log likelihood:
\begin{align*}
\mathcal{L} = -\sum^{N}_{i=1} \sum^{T}_{t=1} \log P_{\Theta}(w_t | \mathbf{h})
\end{align*}
We differentiate with respect to this cost function to calculate gradients.

\subsection{GRU, LSTM and BERT variants}

Does visually situated language learning benefit from the specific architecture of the $\Delta$-RNN, or does the proposal work with state-of-the-art language models?
We applied the same architecture to Gated Recurrent Units \citep[GRU,][]{cho2014learning}, Long Short Term Memory \citep[LSTM,][]{Hochreiter:1997}, and BERT \citep{devlin2018bert}. We train these models on text alone and compare to the two variations of the multi-modal $\Delta$-RNN, as described in the previous section. The multi-modal GRU, with context information directly integrated, is defined as follows:
\begin{align*}
\mathbf{d}_c &= M \mathbf{c} \\
\mathbf{z}_t &= \sigma(W_z \mathbf{x}_t + V_z \mathbf{h}_{t-1})\\
\mathbf{r}_t &= \sigma(W_r \mathbf{x}_t + V_r \mathbf{h}_{t-1})\\
\mathbf{\widehat{h}}_t &= \tanh(W_{\widehat{h}} \mathbf{x}_t + V_{\widehat{h}} (\mathbf{r}_t \otimes \mathbf{h}_{t-1}))\\
\mathbf{h}_t &= [ \mathbf{z}_t \otimes \mathbf{h}_{t-1} + (1 - \mathbf{z}_t) \otimes \mathbf{\widehat{h}}_{t} ] \otimes \mathbf{d}_c
\end{align*}
where we note the parameter matrix $M$ that maps the visual context $c$ into the GRU state effectively gates the outer function.\footnote{We tried both methods of integration, Equations \ref{eqn:inner_integration} and \ref{eqn:outer_integration}. The second formulation gave better performance.} The multi-modal variant of the LSTM (with peephole connections) is defined as follows:
\begin{align*}
\mathbf{d}_c &= M \mathbf{c} \\
\mathbf{h}_{t} &= [ \mathbf{r}_t \otimes \Phi(\mathbf{c}_t) ] \otimes \mathbf{d}_c  \mbox{, where,} \\ 
\mathbf{r}_{t} &= \sigma(W_r \mathbf{x}_t + V_r \mathbf{h}_{t-1} + U_r \mathbf{c}_t) \label{lstm_outer} \\
\mathbf{c}_t &= \mathbf{f}_t \otimes \mathbf{c}_{t-1} + \mathbf{i}_t \otimes \mathbf{z}_t \mbox{, where,} \\
\mathbf{z}_t &= \Phi(W_z \mathbf{x}_t + V_z \mathbf{h}_{t-1}) \mbox{,} \\
\mathbf{i}_t &= \sigma(W_i \mathbf{x}_t + V_i \mathbf{h}_{t-1} + U_i \mathbf{c}_{t-1}) \mbox{,} \\
\mathbf{f}_t &= \sigma(W_f \mathbf{x}_t + V_f \mathbf{h}_{t-1} + U_f \mathbf{c}_{t-1}) \mbox{.}
\end{align*}
We furthermore created one more variant of each multi-modal RNN by initializing a portion of their input-to-hidden weights with embeddings extracted from the Bidirectional Encoder Representations from Transformers (BERT) model \cite{devlin2018bert}. This would correspond to initializing $W$ in the $\Delta$-RNN, $W_i$ in the LSTM, and $W_{\hat{h}}$ in the GRU. Note that in our results, we only report the best-performing model, which turned out to be the LSTM variant.
Since the models in this work are at the word level and BERT operates at the subword level, we create initial word embeddings by first decomposing each word into its appropriate subword components, according to the WordPieces model \cite{wu2016google}, and then extract the relevant BERT representation for each. For each subword token, a representation is created by summing together a specific learned token embedding, a segmentation embedding, and a position embedding. For a target word, we linearly combine subword input representations and initialize the relevant weight with this final embedding.

\section{Experiments}
\label{results}

\begin{figure*}
\centering     
\subfigure[English $\Delta$-RNNs.]
{\label{fig:drnn}\includegraphics[width=0.6\columnwidth]{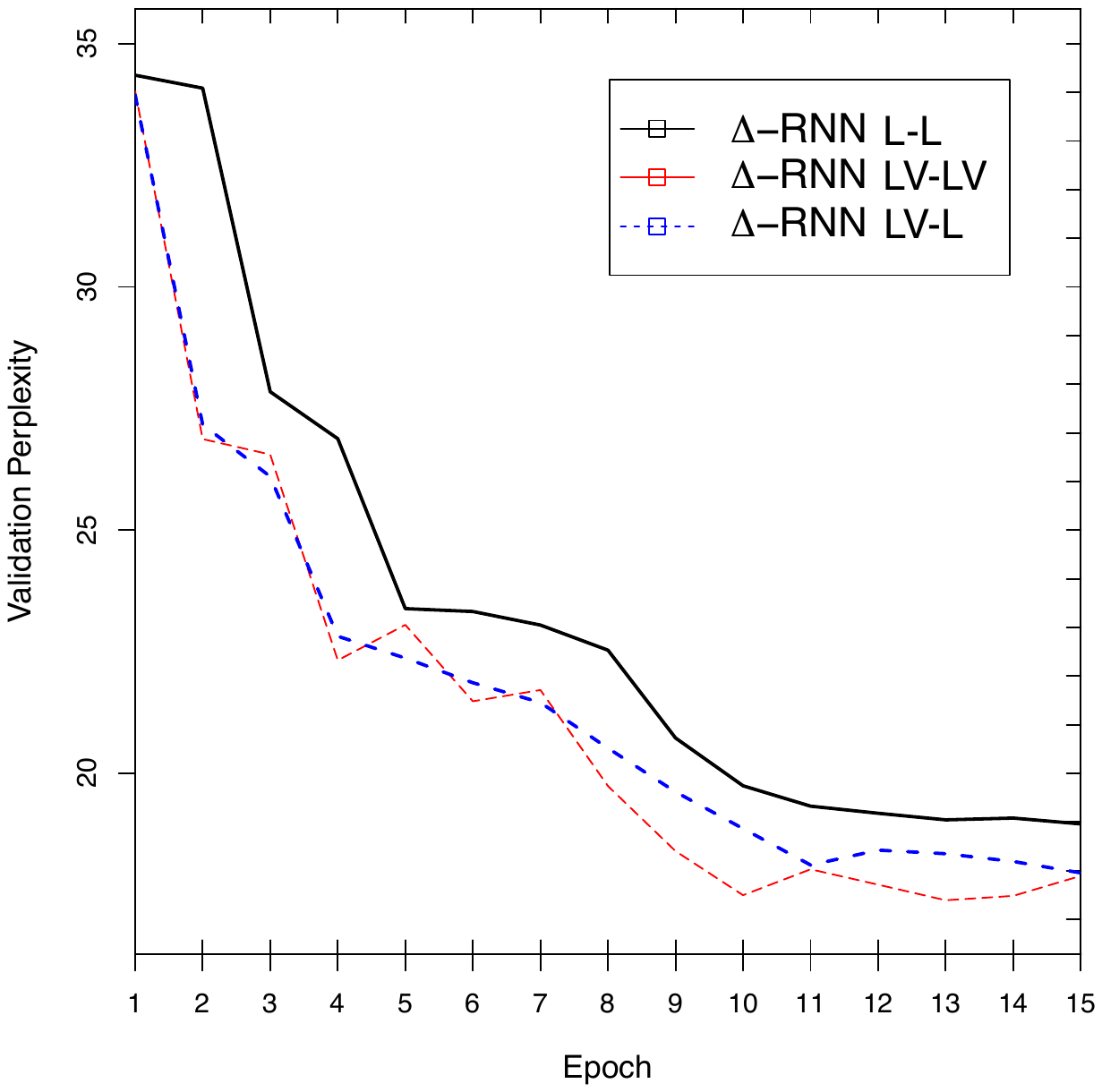}}
\subfigure[German $\Delta$-RNNs.]
{\label{fig:mmdrnn}\includegraphics[width=0.6\columnwidth]{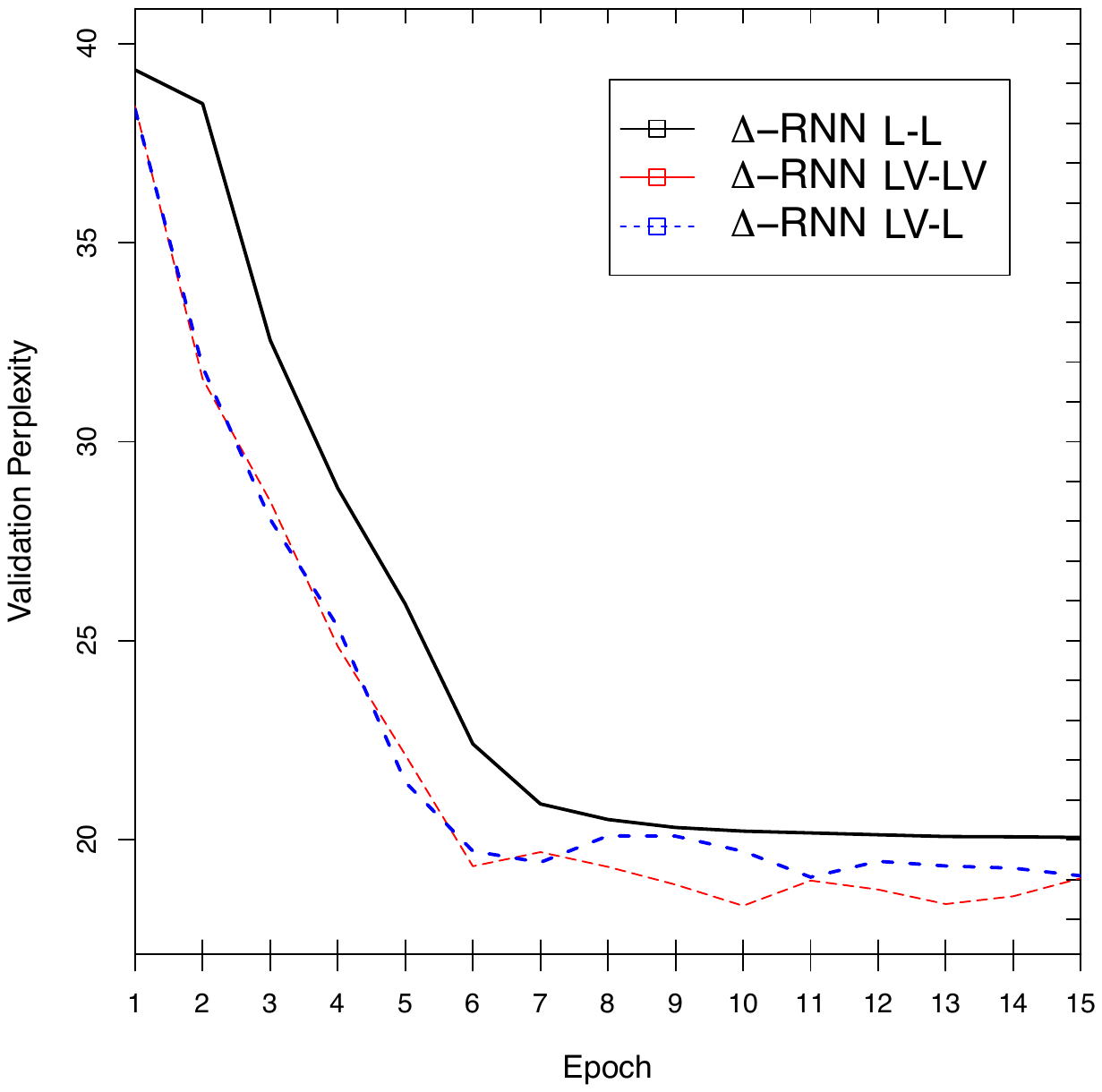}}
\subfigure[Spanish $\Delta$-RNNs.]
{\label{fig:mmdrnn_impov}\includegraphics[width=0.6\columnwidth]{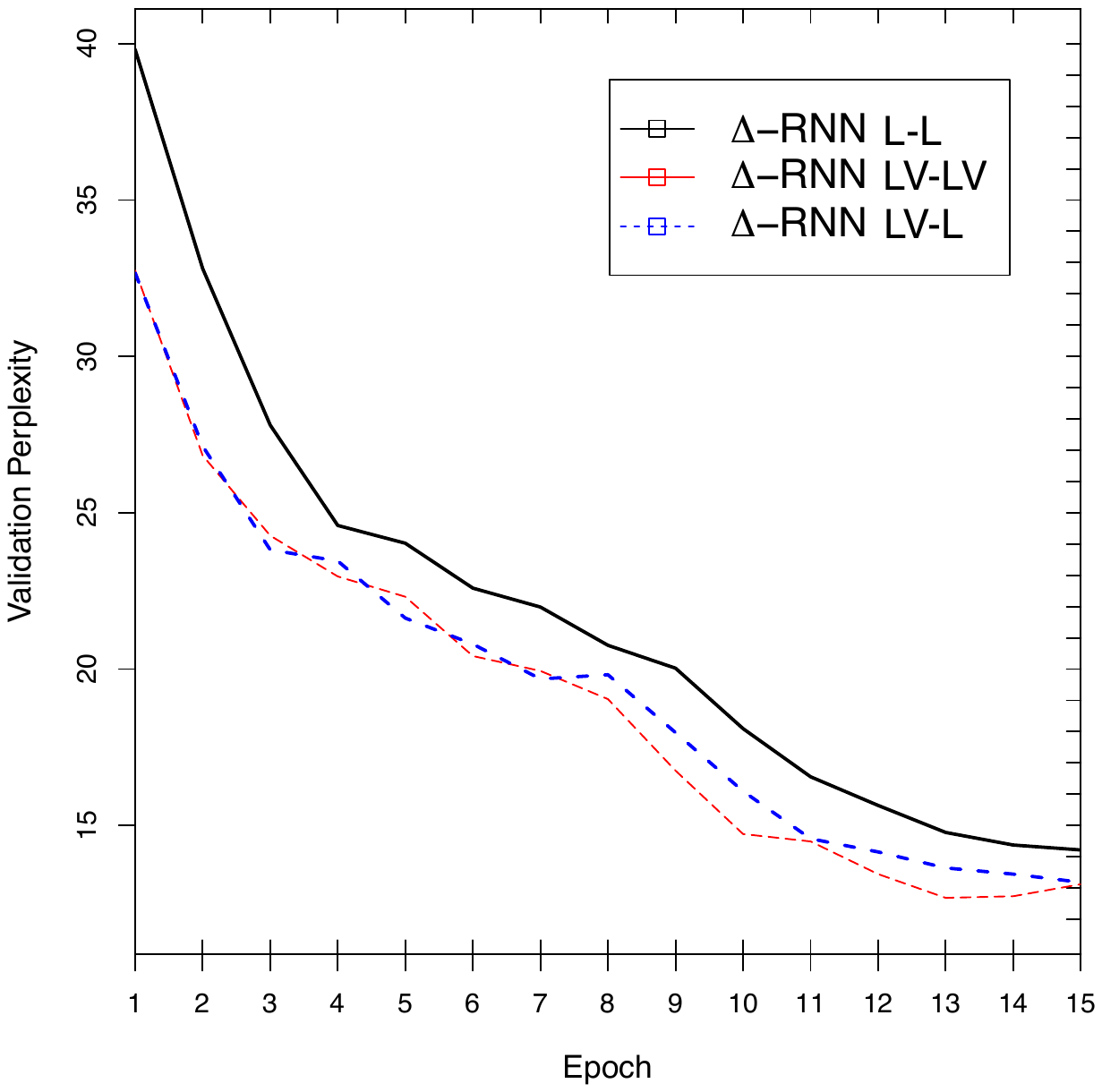}}
\caption{Training $\Delta$-RNNs in each language (English, German, Spanish). Baseline model is trained and evaluated on language (L-L), the \emph{full} model uses the multi-modal signal (LV-LV), and the target model is trained on LV, but evaluated on L only (LV-L).}
\label{fig:learning_curves}
\end{figure*}

The experiments in this paper were conducted using the MS-COCO image-captioning dataset.\footnote{https://competitions.codalab.org/competitions/3221} Each image in the dataset has five captions provided by human annotators. We use the captions to create five different ground truth splits. We translated each ground truth split into German and Spanish using the Google Translation API, which was chosen as a state-of-the-art, independently evaluated MT tool that produces, according to our inspection of the results, idiomatic, and syntactically and semantically faithful translations. To our knowledge, this represents the first  Multi-lingual MSCOCO dataset on situated learning.  
We tokenize the corpus and obtain a 16.6K vocabulary for English, 33.2K for German and 18.2k for Spanish.

\input{model-performance-table}

As our primary concern is the next-step prediction of words/tokens, we use negative log likelihood and perplexity to evaluate the models. This is different from the goals of machine translation or image captioning, which, in most cases, is concerned with a ranking of possible captions where one measures how similar the model's generated sequences are to ground-truth target phrases.

Baseline results were obtained with neural language models trained on text alone. For the $\Delta$-RNN, this meant implementing a model using only Equations 1-7. The best results were achieved using the BERT Large model (bidirectional Transformer, 24 layers, 1024dims, 16 attention heads: \citealt{devlin2018bert}). We used the large pretrained model and then trained with visual context.

All models were trained to minimize the sequence loss of the sentences in the training split. The weight matrices of all models were initialized from uniform distribution, $U(-0.1, 0.1)$, biases were initialized from zero, and the $\Delta$-RNN-specific biases $\{\alpha, \beta_1, \beta_2\}$ were all initialized to one. Parameter updates calculated through back-propagation through time required unrolling the model over 49 steps in time (this length was determined based on validation set likelihood). All symbol sequences were zero-padded and appropriately masked to ensure efficient mini-batching. Gradients were hard-clipped at a magnitude bound of $l = 2.0$. Over mini-batches of 32 samples, model parameters were optimized using simple stochastic gradient descent (learning rate $\lambda = 1.0$ which was halved if the perplexity, measured at the end of each epoch, goes up three or more times).


\input{conditional-sampling.tex}
\input{word-query.tex}
To determine if our multi-modal language models capture knowledge that is different from a text-only language model, we evaluate each model twice. First, we compute the model perplexity on the test set using the sentences' visual context vectors. Next, we compute model perplexity on test sentences by feeding in a null-vector to the multi-modal model as the visual context. If the model did truly pick up some semantic knowledge that is not exclusively dependent on the context vector, its perplexity in the second setting, while naturally worse than the first setting, should still outperform text-only baselines. 

In Table \ref{results:generalization}, we report each model's negative log likelihood (NLL) and per-word perplexity (PPL). PPL is calculated as:
\begin{align*}
PPL = \exp \big [-(1/N) \sum^{N}_{i=1} \sum^{T}_{t=1} \log P_{\Theta}(w_t | \mathbf{h}) \big ]
\end{align*}
We observe that in all cases the multi-modal models outperform their respective text-only baselines. More importantly, the multi-modal models, when evaluated without the Inception-v3 representations on holdout samples, still perform better than the text-only baselines. The improvement in language generalization can be attributed to the visual context information provided during training, enriching its representations over word sequences with knowledge of actual objects and actions.

Figure \ref{fig:learning_curves} shows the validation perplexity of the $\Delta$-RNN on each language as a function of the first 15 epochs of learning. We observe that throughout learning, the improvement in generalization afforded by the visual context $\mathbf{c}$ is persistent. Validation performance was also tracked for the various GRU and LSTM models, where the same trend was also observed (see supplementary material). 

\subsection{Model Analysis}
\label{analysis}
We analyze the decoders of text-only and multi-modal models. We examine the parameter matrix $U$, which is directly involved in calculating the predictions of the underlying generative model. $U$ can be thought of as ``transposed embeddings'', an idea that has also been exploited to introduce further regularization into the neural language model learning process \citep{press2016using,inan2016tying}. If we treat each row of this matrix as the learned embedding for a particular word (we assume column-major orientation in implementation), we can calculate its proximity to other embeddings using cosine similarity.

Table \ref{results:word_query} shows the top ten words for several randomly selected query terms using the decoder parameter matrix. By observing the different sets of nearest-neighbors produced by the $\Delta$-RNN and the multi-modal $\Delta$-RNN (MM-$\Delta$-RNN), we can see that the MM-$\Delta$-RNN appears to have learned to combine the information from the visual context with the token sequence in its representations. For example, for the query ``ocean'', we see that while the $\Delta$-RNN does associate some relevant terms, such as ``surfing'' and ``beach'', it also associates terms with marginal relevance to ``ocean'' such as ``market'' and ``plays''. Conversely, nearly all of the terms the MM-$\Delta$-RNN associates with ``ocean'' are relevant to the query. The same is true for ``kite'' and ``subway''. For ``racket'', while the text-only baseline mostly associates the query with sports terms, especially sports equipment like ``bat'', the MM-$\Delta$-RNN is able to relate the query to the correct sport, ``tennis''.


\subsection{Conditional Sampling}
\label{exp:samples}
To see how visual context influences the language model, we sample the conditional generative model. 
Beam search (size $13$) allows us to generate full sentences (Table~\ref{results:samples}). Words were ranked based on model probabilities.

\section{Discussion and Conclusions}
\label{conclude}

Training with perceptual context improves  multi-modal neural models compared to training on language alone.
Specifically, augmenting a predictive language model with images that illustrate the sentences being learned enhances its next-word or masked-word prediction ability. The performance improvement persists even in situations devoid of visual input, when the model is  used as a pure language model.

The near state-of-the-art language model, using BERT, reflects the case of human language acquisition less than do the other models, which were trained ``ab initio" in a situated context.  BERT is pre-trained on a very large corpus, but it still picked up a performance improvement when fine-tuned on the visual context and language, as compared to the corpus language signal alone.  We do not expect this to be a ceiling for visual augmentation: in the world of training LMs, the MS COCO corpus is, of course, a small dataset.

Neural language models, as used here, are contenders as cognitive and psycholinguistic models of the non-symbolic, implicit aspects of language representation. There is a great deal of evidence that something like a predictive language model exists in the human mind. The \emph{surprisal} of a word or phrase refers to the degree of mismatch between what a human listener expected to be said next and what is actually said, for example, when a garden path sentence forces the listener to abandon a partial, incremental parse \citep{ferreira1991recovery,hale2001probabilistic}. In the garden path sentence ``The horse raced past the barn fell'', the final word ``fell'' forces the reader to revise their initial interpretation of ``raced'' as the active verb \cite{bever1970cognitive}.  

More generally, the idea of \emph{predictive coding} holds that the mind forms expectations before perception occurs \cite[see][for a review]{clark2013whatever}. How these predictions are formed is unclear. Predictive language models trained with a generic neural architecture, without specific linguistic universals, are a reasonable candidate for a model of predictive coding in language.  This does not imply neuropsychological realism of the low-level representations or learning algorithms, and we cannot advocate for a specific neural architecture as being most plausible.  However, we can show that an architecture that predicts linguistic input well learns better when its input mimics that of a human language learner.  

A theory of human language processing might distinguish between symbolic language knowledge and processes that implement compositionality to produce semantics on the one hand, and implicit processes that leverage sequences and associations to produce expectations.  With respect to acquiring the latter, implicit and predictive model, we note that children are exposed to a rich sensory environment, one more detailed than what is provided to our model here.  If even static visual input alone improves language acquisition, then what could a sensorily rich environment achieve?  When a multi-modal learner is considered, then, perhaps, the language acquisition stimulus that has been famously labeled to be rather poor \citep{chomsky1959review,berwick2013poverty}, is quite rich after all.



\section*{Acknowledgments}
We would like to thank Tomas Mikolov, Emily Pitler, Zixin Tang, and Saranya Venkatraman for comments. Part of this work was funded by the National Science Foundation (BCS-1734304 to D. Reitter).

\bibliography{visuallm}
\bibliographystyle{acl_natbib}

\input{appendix}

\end{document}

%% file: model-performance-table.tex
\begin{table*}[t]
\renewcommand{\arraystretch}{1.6}
\begin{center}

\footnotesize

\sisetup{detect-weight,mode=text}
\renewrobustcmd{\boldmath}{}
\newrobustcmd{\B}{\bfseries}

\begin{tabular}{ @{} l *{6}{S[table-format=1.3]} @{}}
  & \multicolumn{2}{c}{\textbf{English}} & \multicolumn{2}{c}{\textbf{German MT}} & \multicolumn{2}{c}{\textbf{Spanish MT}} \\
  Model (Type) & {Test-NLL} & {Test-PPL} & {Test-NLL} & {Test-PPL} & {Test-NLL} & {Test-PPL}\\
  \hline
  $\Delta$-RNN (L-L)& 2.714 & 15.086 & 2.836 & 17.052 & 2.546 & 12.755 \\
  MM-$\Delta$-RNN (LV-LV) & 2.645 & 14.086 & 2.777 & 16.082 & 2.405 & 11.082 \\
  MM-$\Delta$-RNN (LV-L) & 2.694 & 14.786 & 2.808 & 16.582 & 2.458 & 11.682 \\
  \hline
  GRU (L-L)& 2.764 & 15.871 & 2.854 & 17.369 & 2.554 & 12.866 \\ 
  MM-GRU (LV-LV) & 2.654 & 14.189 & 2.790 & 16.285 & 2.426 & 11.3089 \\
  MM-GRU (LV-L) & 2.687 & 14.689 & 2.815 & 16.701 & 2.466 & 11.781 \\
  \hline
  LSTM (L-L)& 2.722 & 15.217 & 2.814 & 17.070 & 2.494 & 12.114 \\
  MM-LSTM (LV-LV) & 2.645 & 14.089 & 2.773 & 16.001 & 2.405 & 11.081 \\
  MM-LSTM (LV-L) & 2.708 & 15.002 & 2.822 & 16.806 & 2.487 & 12.028 \\
  \hline
  BERT+LSTM (L-L)& 2.534 & 12.6011 & 2.702 & 14.9127 & 2.303 & 10.0011 \\
  BERT+MM-LSTM (LV-LV) & 2.475 & 11.8776 & 2.661 & 14.3124 & 2.223 & 9.2319 \\
  BERT+MM-LSTM (LV-L) & \B 2.503 & \B 12.2196 & \B 2.700 & \B 14.8102 & \B 2.283 & \B 9.8102 \\
\end{tabular}
\caption{Generalization performance as measured by negative log likelihood (NLL) and perplexity (PPL). Lower values indicate better performance. Baseline model (L-L) trained and evaluated on linguistic data only. Full model (LV-LV) trained and evaluated on both linguistic and visual data. Blind model (LV-L) trained on both but evaluated on language only. \textbf{The difference between L-L and LV-L illustrates the performance improvement.} German and Spanish data are machine-translated (MT) and provide additional, but correlated, evidence.  For comparison, \citet{devlin2018bert} report a perplexity of $3.23$ for their (broad) English test data, using the same base model we use here to define input representations.}
\label{results:generalization}
\end{center}
\end{table*}

%% file: conditional-sampling.tex
\noindent\begin{table}[t]\renewcommand{\arraystretch}{1.0}%
\footnotesize%
\begin{tabular}[]{l l}%
\includegraphics[width=0.2\columnwidth]{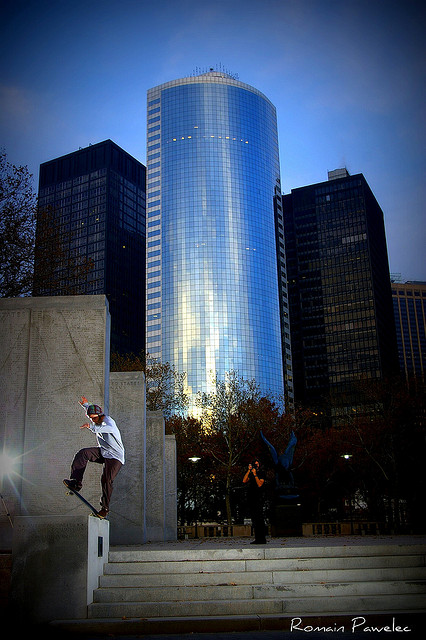}&
\begin{minipage}{0.6\columnwidth}%
\vskip-1.8cm%
\hangindent=2em a skateboarder and person in front of skyscrapers.

\hangindent=2em a person with skateboarder on air.

\hangindent=2em a person doing a trick with skateboarder.

\hangindent=2em a person with camera with blue background.\\ 
\end{minipage}
\\
\hline\\
\includegraphics[width=0.2\columnwidth]{figs/COCO_2.jpg}& 
\begin{minipage}{0.6\columnwidth}%
\vskip-1.0cm%
a food bowl on the table \\ a bowl full of food on the table

\hangindent=2em a green and red bowl on the table

a salad bowl with chicken\\
\end{minipage}\\
\hline\\
\includegraphics[width=0.2\columnwidth]{figs/COCO_3.jpg}&
\begin{minipage}{0.6\columnwidth}
\vskip-0.7cm%
a dog on blue bed with blanket. \\ a dog sleeps near wooden table. \\ a dog sleeps on a bed. \\ a dog on some blue blankets.\\
\end{minipage}
\end{tabular}
\caption{Some captions generated by the multi-modal $\Delta$-RNN in English.}
\label{results:samples}%
\end{table}

%% file: word-query.tex
\begin{table*}
\renewcommand{\arraystretch}{1.4}
\centering
\small
\begin{tabular}{l|l||l|l||l|l||l|l}

\multicolumn{2}{l||}{\textbf{Ocean}} & \multicolumn{2}{l|}{\textbf{Kite}} & \multicolumn{2}{l|}{\textbf{Subway}} & \multicolumn{2}{l}{\textbf{Racket}}\\
\hline
$\Delta$-\textbf{RNN} & +MM & $\Delta$-\textbf{RNN} & +MM & $\Delta$-\textbf{RNN} & +MM & $\Delta$-\textbf{RNN} & +MM\\
\hline
surfing & boats & plane & kites & train & railroad & bat & bat \tabularnewline
sandy & beach & kites & airplane & passenger & train & batter & players \tabularnewline
filled & pier & airplane & plane & railroad & locomotive & catcher & batter \tabularnewline
beach & wetsuit & surfboard & airplanes & trains & trains & skateboard & swing \tabularnewline
market & cloth & planes & planes & gas & steam & umpire & catcher \tabularnewline
crowded & surfing & airplanes & airliner & commuter & gas & soccer & hitter \tabularnewline
topped & windsurfing& boats & helicopter & trolley & commuter & women & ball \tabularnewline
plays & boardwalk & jet & jets & locomotive & passenger & pedestrians & umpire \tabularnewline
cross & flying & aircraft & biplane & steam & crowded & players & tennis \tabularnewline
snowy & biplane & jets & jet & it's & trolley & uniform & tatoos \tabularnewline
\end{tabular}%
\caption{The ten words most closely related to the bolded query word, rank ordered, trained without ($\Delta$-RNN) and with (+MM) visual input.}%
\label{results:word_query}%
\end{table*}

%% file: appendix.tex
\clearpage
\appendix
\begin{figure*}[h!]
\subfigure[English GRUs.]
{\label{fig:gru}\includegraphics[width=50mm]{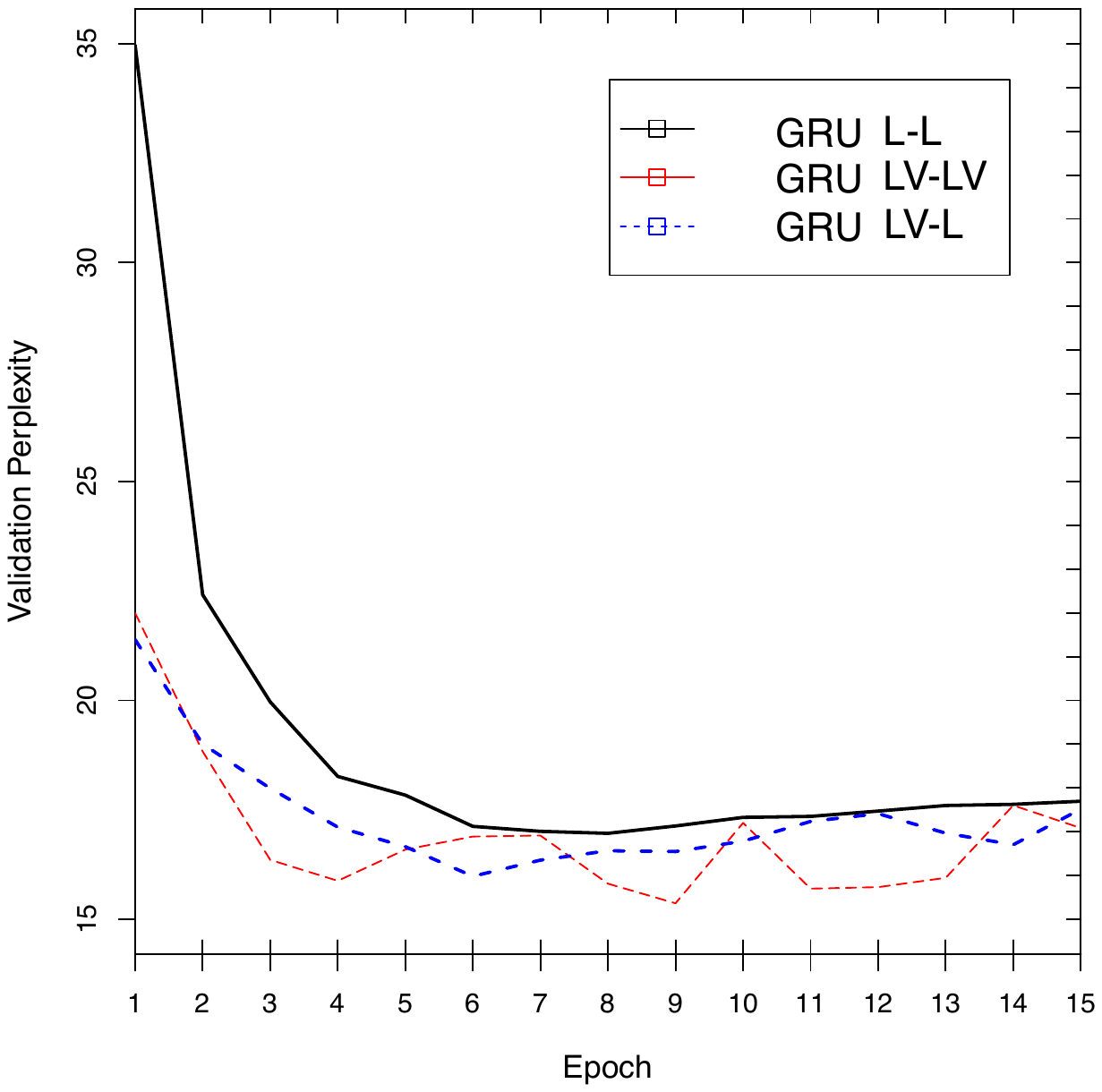}}
\subfigure[German GRUs.]
{\label{fig:mmgru}\includegraphics[width=50mm]{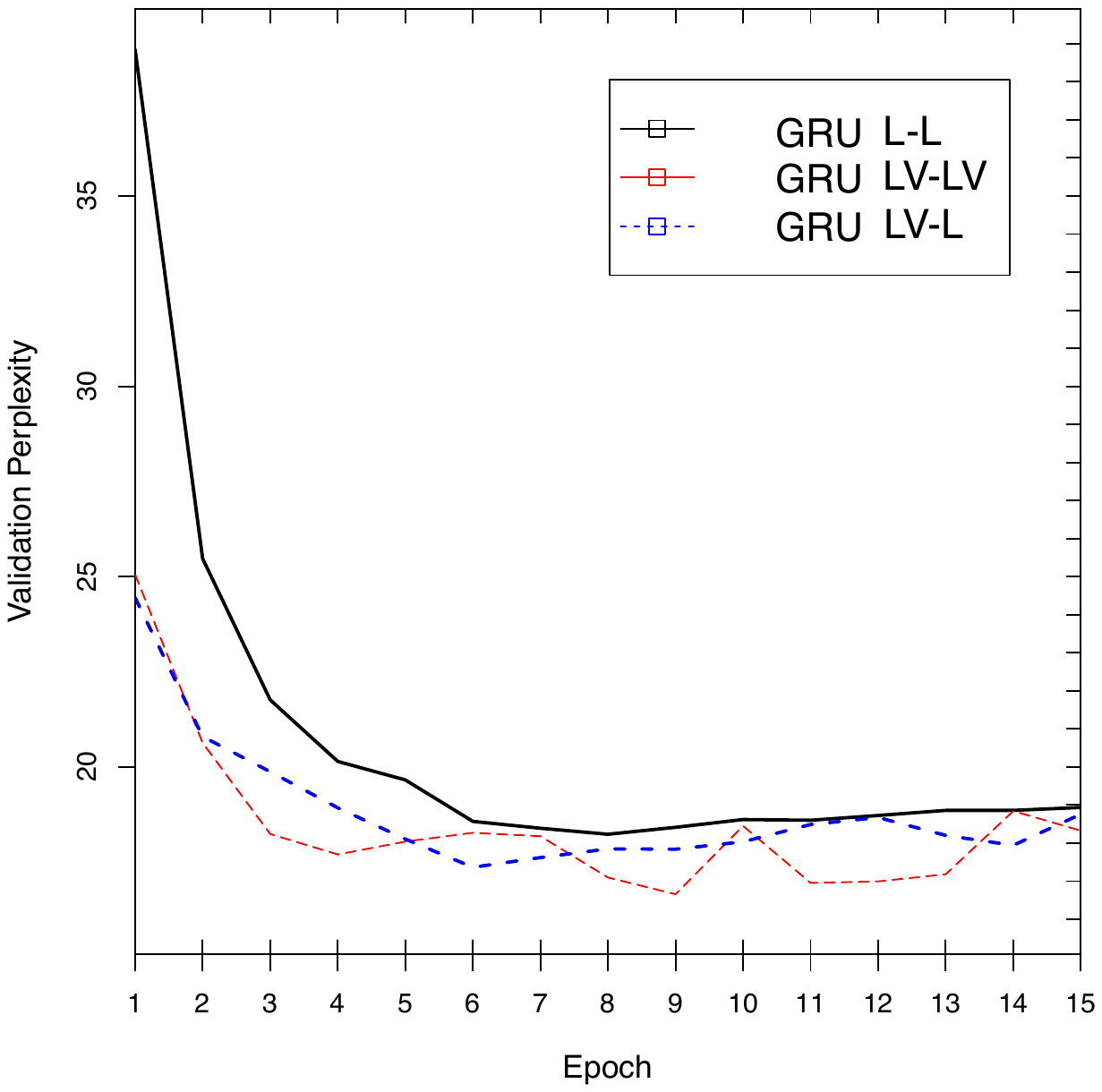}}
\subfigure[Spanish GRUs.]
{\label{fig:mmgru_impov}\includegraphics[width=50mm]{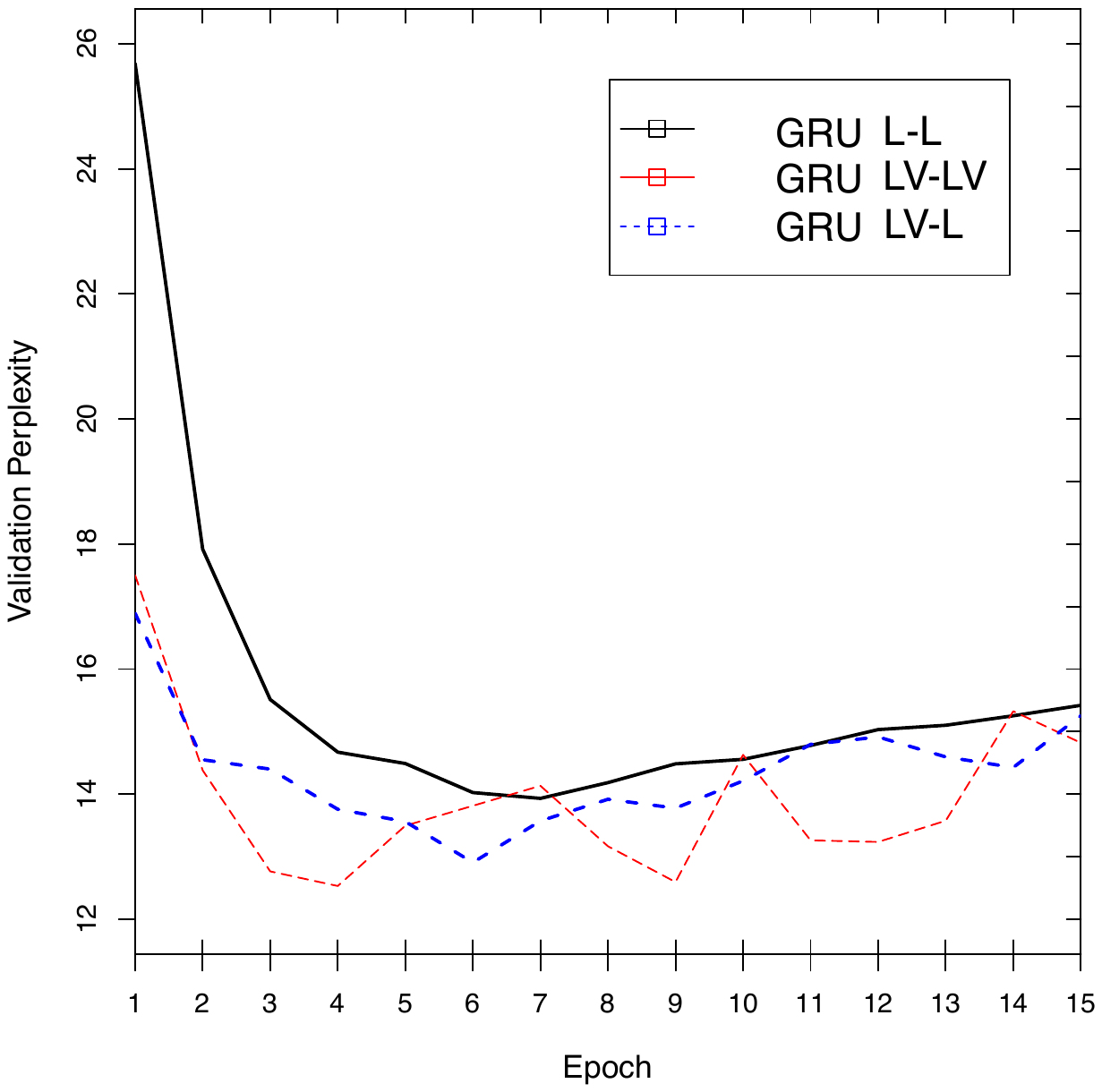}}%

\subfigure[English LSTMs.]
{\label{fig:lstm}\includegraphics[width=50mm]{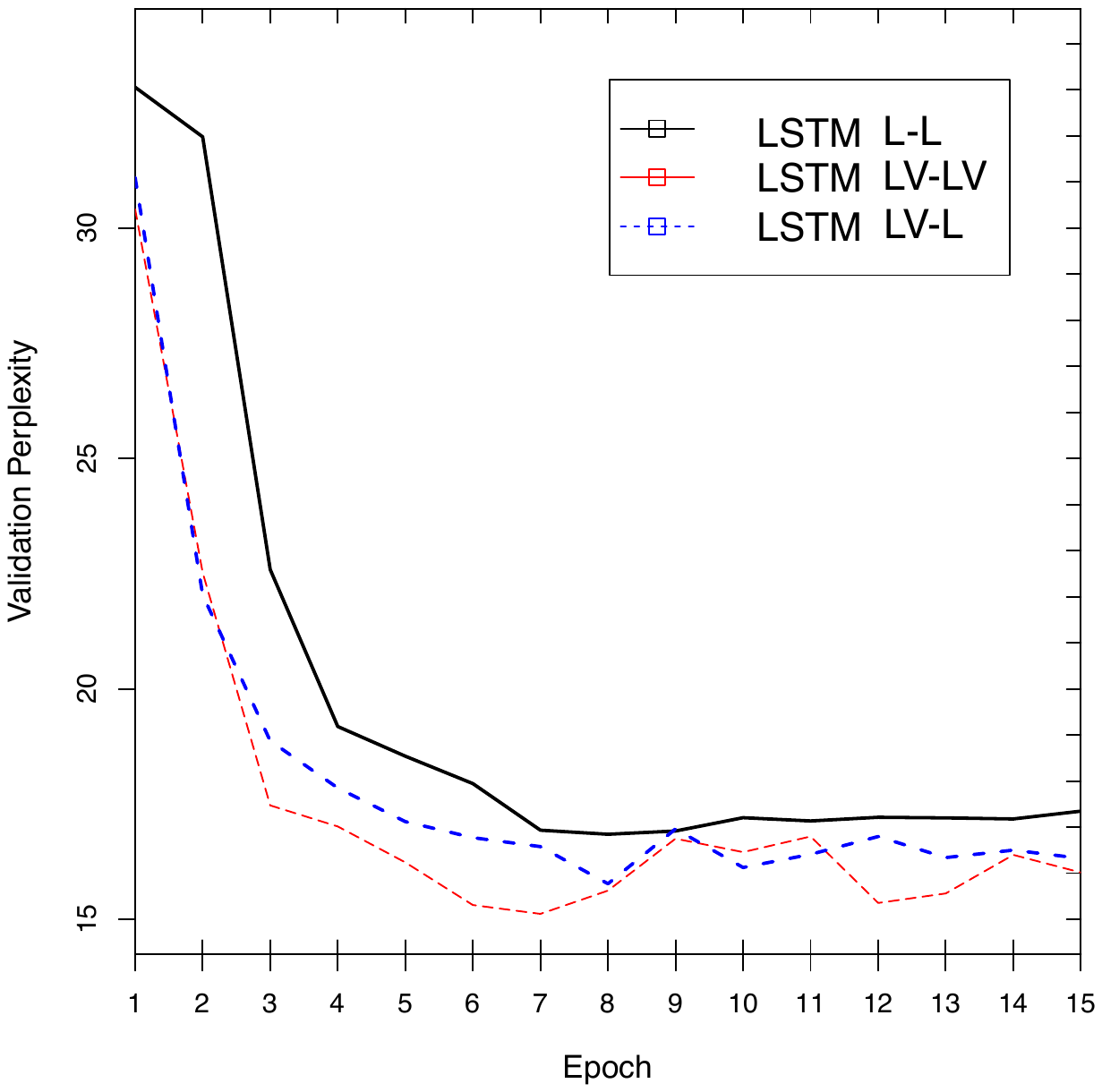}}
\subfigure[German LSTMs.]
{\label{fig:mmlstm}\includegraphics[width=50mm]{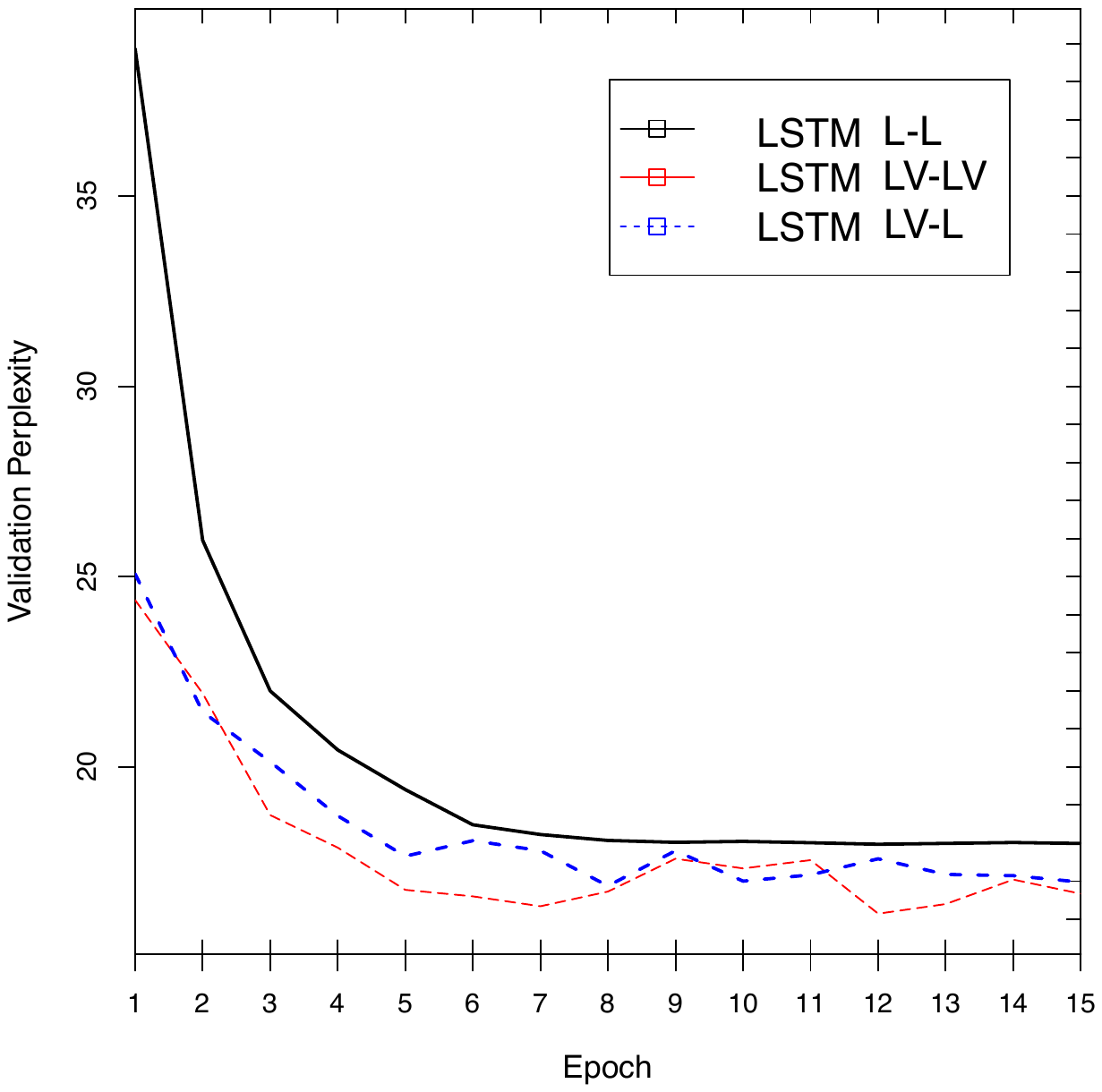}}
\subfigure[Spanish LSTMs.]
{\label{fig:mmlstm_impov}\includegraphics[width=50mm]{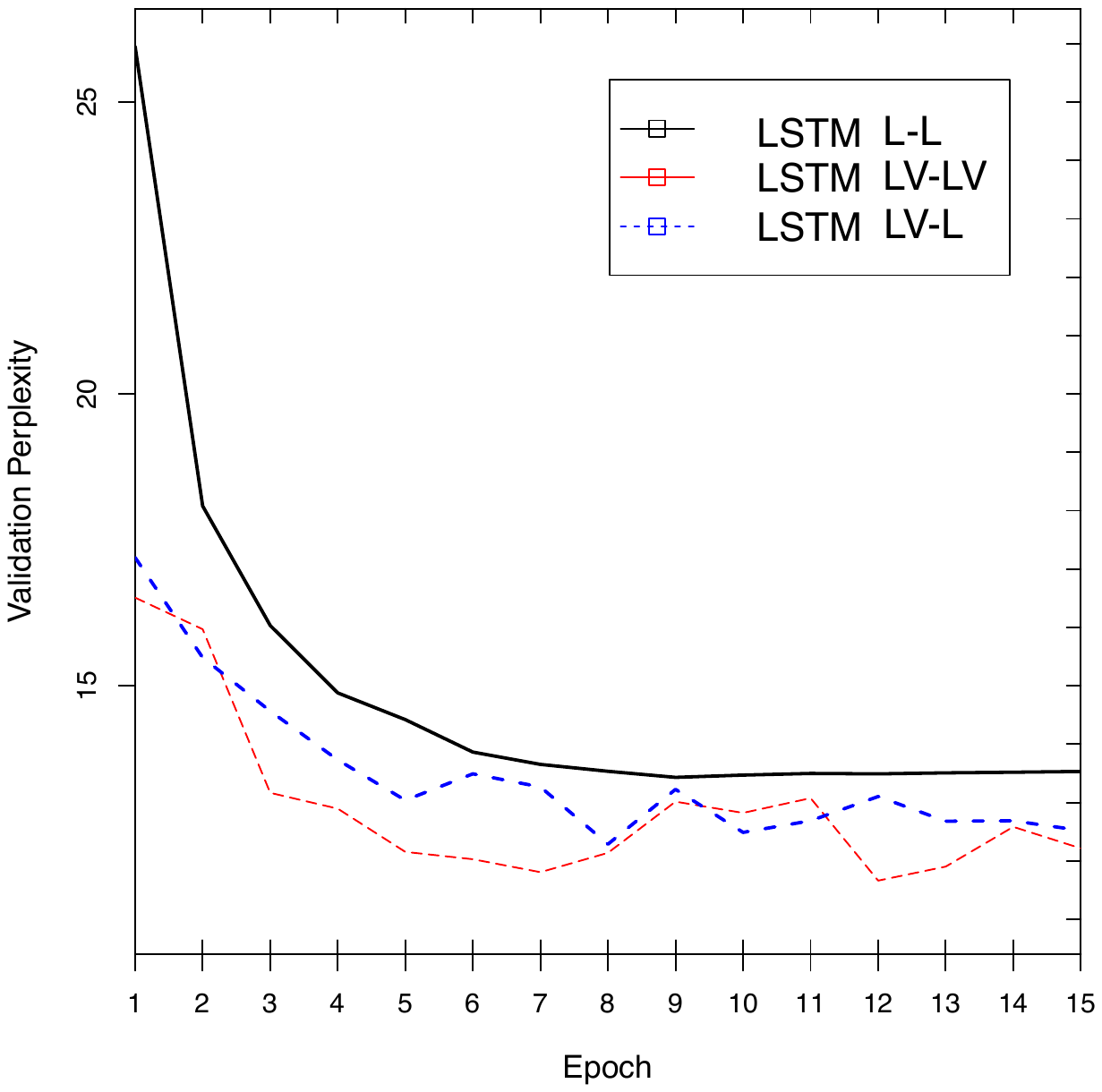}}

\caption*{\textbf{Appendix:}  Comparison of learning curves for the GRUs and LSTMs in each language (English, German, Spanish). To augment Figure 2 in the main paper, we also show the learning curves for all models experimented with in this paper beyond the $\Delta$-RNN. Validation learning curves are provided for the GRU and LSTM language models, both multimodal and unimodal variations. }
\label{fig:models}
\end{figure*}

%% file: visuallm.bib
@article{ferreira1991recovery,
  title={Recovery from misanalyses of garden-path sentences},
  author={Ferreira, Fernanda and Henderson, John M},
  journal={Journal of Memory and Language},
  volume={30},
  number={6},
  pages={725--745},
  year={1991},
  publisher={Elsevier}
}

@inproceedings{bever1970cognitive,
  title={The cognitive basis for linguistic structures},
  author={Bever, Thomas G},
  booktitle={Cognition and the development of language},
  editor={J. R. Hayes},
  pages={279-362},
  year={1970},
}

@article {Tanenhaus1995,
	author = {Tanenhaus, MK and Spivey-Knowlton, MJ and Eberhard, KM and Sedivy, JC},
	title = {Integration of visual and linguistic information in spoken language comprehension},
	volume = {268},
	number = {5217},
	pages = {1632--1634},
	year = {1995},
	doi = {10.1126/science.7777863},
	publisher = {American Association for the Advancement of Science},
	issn = {0036-8075},
	URL = {https://science.sciencemag.org/content/268/5217/1632},
	eprint = {https://science.sciencemag.org/content/268/5217/1632.full.pdf},
	journal = {Science}
}

@article{Baroni2016visual,
author = {Baroni, Marco},
title = {Grounding Distributional Semantics in the Visual World},
journal = {Language and Linguistics Compass},
volume = {10},
number = {1},
pages = {3-13},
doi = {10.1111/lnc3.12170},
url = {https://onlinelibrary.wiley.com/doi/abs/10.1111/lnc3.12170},
eprint = {https://onlinelibrary.wiley.com/doi/pdf/10.1111/lnc3.12170},
year = {2016}
}

@inproceedings{lazaridou2014wampimuk,
  title={Is this a wampimuk? cross-modal mapping between distributional semantics and the visual world},
  author={Lazaridou, Angeliki and Bruni, Elia and Baroni, Marco},
  booktitle={Proceedings of the 52nd Annual Meeting of the Association for Computational Linguistics (Volume 1: Long Papers)},
  volume={1},
  pages={1403--1414},
  year={2014}
}

@article{Lazaridou2015adjectives,
author = {Lazaridou, Angeliki and Dinu, Georgiana and Liska, Adam and Baroni, Marco},
title = {From Visual Attributes to Adjectives through Decompositional Distributional Semantics},
journal = {Transactions of the Association for Computational Linguistics},
volume = {3},
number = {},
pages = {183-196},
year = {2015},
doi = {10.1162/tacl\_a\_00132},
URL = {https://doi.org/10.1162/tacl_a_00132},
}

@inproceedings{lazaridou2015multimodal,
    title = "Combining Language and Vision with a Multimodal Skip-gram Model",
    author = "Lazaridou, Angeliki  and
      Pham, Nghia The  and
      Baroni, Marco",
    booktitle = "Proceedings of the 2015 Conference of the North {A}merican Chapter of the Association for Computational Linguistics: Human Language Technologies",
    month = may # "{--}" # jun,
    year = "2015",
    address = "Denver, Colorado",
    publisher = "Association for Computational Linguistics",
    url = "https://www.aclweb.org/anthology/N15-1016",
    doi = "10.3115/v1/N15-1016",
    pages = "153--163",
}

@article{devlin2018bert,
  title={{BERT}: Pre-training of deep bidirectional transformers for language understanding},
  author={Devlin, Jacob and Chang, Ming-Wei and Lee, Kenton and Toutanova, Kristina},
  journal={arXiv preprint arXiv:1810.04805},
  year={2018}
}

@article{clark2013whatever,
  title={Whatever next? {P}redictive brains, situated agents, and the future of cognitive science},
  author={Clark, Andy},
  journal={Behavioral and brain sciences},
  volume={36},
  number={3},
  pages={181--204},
  year={2013},
  publisher={Cambridge University Press}
}

@article{wu2016google,
  title={Google's neural machine translation system: Bridging the gap between human and machine translation},
  author={Wu, Yonghui and Schuster, Mike and Chen, Zhifeng and Le, Quoc V and Norouzi, Mohammad and Macherey, Wolfgang and Krikun, Maxim and Cao, Yuan and Gao, Qin and Macherey, Klaus and others},
  journal={arXiv preprint arXiv:1609.08144},
  year={2016}
}

@article{rao1999predictive,
  title={Predictive coding in the visual cortex: {A} functional interpretation of some extra-classical receptive-field effects},
  author={Rao, Rajesh PN and Ballard, Dana H},
  journal={Nature Neuroscience},
  volume={2},
  number={1},
  pages={79},
  year={1999},
  publisher={Nature Publishing Group}
}

@article{chomsky1959review,
  title={A review of {BF} {S}kinner's Verbal Behavior},
  author={Chomsky, Noam},
  journal={Language},
  volume={35},
  number={1},
  pages={26--58},
  year={1959}
}

@incollection{berwick2013poverty,
  title={Poverty of the stimulus stands: Why recent challenges fail},
  author={Berwick, Robert C and Chomsky, Noam and Piattelli-Palmarini, Massimo},
  booktitle={Rich Languages From Poor Inputs},
  pages={19--42},
  year={2013},
  publisher={Oxford University Press},
  chapter={1},
}

@inproceedings{hale2001probabilistic,
  title={A probabilistic {E}arley parser as a psycholinguistic model},
  author={Hale, John},
  booktitle={Proceedings of the Second Meeting of the North American Chapter of the Association for Computational Linguistics on Language technologies},
  pages={1--8},
  year={2001},
  address={Pittsburgh, PA},
  _organization={Association for Computational Linguistics}
}

@article{press2016using,
  title={Using the output embedding to improve language models},
  author={Press, Ofir and Wolf, Lior},
  journal={arXiv preprint arXiv:1608.05859},
  year={2016}
}

@article{inan2016tying,
  title={Tying word vectors and word classifiers: A loss framework for language modeling},
  author={Inan, Hakan and Khosravi, Khashayar and Socher, Richard},
  journal={arXiv preprint arXiv:1611.01462},
  year={2016}
}

@article{socher2014grounded,
  title={Grounded compositional semantics for finding and describing images with sentences},
  author={Socher, Richard and Karpathy, Andrej and Le, Quoc V and Manning, Christopher D and Ng, Andrew Y},
  journal={Transactions of the Association of Computational Linguistics},
  volume={2},
  number={1},
  pages={207--218},
  year={2014}
}

@article{kiros2014unifying,
  title={Unifying visual-semantic embeddings with multimodal neural language models},
  author={Kiros, Ryan and Salakhutdinov, Ruslan and Zemel, Richard S},
  journal={arXiv preprint arXiv:1411.2539},
  year={2014}
}

@inproceedings{frome2013devise,
  title={Devise: A deep visual-semantic embedding model},
  author={Frome, Andrea and Corrado, Greg S and Shlens, Jon and Bengio, Samy and Dean, Jeff and Mikolov, Tomas and others},
  booktitle={Advances in neural information processing systems},
  pages={2121--2129},
  year={2013}
}

@inproceedings{szegedy2016rethinking,
  title={Rethinking the inception architecture for computer vision},
  author={Szegedy, Christian and Vanhoucke, Vincent and Ioffe, Sergey and Shlens, Jon and Wojna, Zbigniew},
  booktitle={Proceedings of the IEEE Conference on Computer Vision and Pattern Recognition},
  pages={2818--2826},
  year={2016}
}

@article{barsalou1999perceptions,
  title={Perceptions of perceptual symbols},
  author={Barsalou, Lawrence W},
  journal={Behavioral and Brain Sciences},
  volume={22},
  number={4},
  pages={637--660},
  year={1999},
  publisher={Cambridge University Press}
}

@article{barsalou2008grounded,
  title={Grounded cognition},
  author={Barsalou, Lawrence W},
  journal={Annual Review of Psychology},
  volume={59},
  pages={617--645},
  year={2008},
  publisher={Annual Reviews}
}

@inproceedings{alishahi2008fast,
  title={Fast mapping in word learning: What probabilities tell us},
  author={Alishahi, Afra and Fazly, Afsaneh and Stevenson, Suzanne},
  booktitle={Proceedings of the Twelfth Conference on Computational Natural Language Learning},
  pages={57--64},
  year={2008},
  organization={Association for Computational Linguistics}
}

@article{abend2017bootstrapping,
	title = {Bootstrapping language acquisition},
	volume = {164},
	issn = {0010-0277},
	url = {http://www.sciencedirect.com/science/article/pii/S0010027717300495},
	doi = {10.1016/j.cognition.2017.02.009},
	journal = {Cognition},
	author = {Abend, Omri and Kwiatkowski, Tom and Smith, Nathaniel J. and Goldwater, Sharon and Steedman, Mark},
	year = {2017},
	pages = {116 -- 143}
}

@inproceedings{frank2008bayesian, title={A {B}ayesian framework for cross-situational word-learning}, author={Frank, Michael C and Goodman, Noah D and Tenenbaum, Joshua B}, booktitle={Advances in neural information processing systems}, pages={457--464}, year={2008} }

@article{greeno1993situativity,
  title={Situativity and symbols: Response to {V}era and {S}imon},
  author={Greeno, James G and Moore, Joyce L},
  journal={Cognitive Science},
  volume={17},
  number={1},
  pages={49--59},
  year={1993},
  publisher={Wiley Online Library}
}

@article{biederman1987recognition,
  title={Recognition-by-components: A theory of human image understanding.},
  author={Biederman, Irving},
  journal={Psychological Review},
  volume={94},
  number={2},
  pages={115},
  year={1987},
  publisher={US: American Psychological Association}
}

@inproceedings{kievit2011semantic,
  title={The semantic Pictionary project},
  author={Kievit-Kylar, Brent and Jones, Michael},
  booktitle = {Proceedings of the 33rd Annual Conference of the Cognitive Science Society},
  year={2011},
  address = {Austin, TX},
  publisher = {Cognitive Science Society},
  editor={L. Carlson, C. Hoelscher, and T.F. Shipley},
  pages={2229--2234},
  url={https://mindmodeling.org/cogsci2011/papers/0520/},
}

@article{johns2012perceptual,
  title={Perceptual inference through global lexical similarity},
  author={Johns, Brendan T and Jones, Michael N},
  journal={Topics in Cognitive Science},
  volume={4},
  number={1},
  pages={103--120},
  year={2012},
  publisher={Wiley Online Library}
}

@article{roy2005connecting,
  title={Connecting language to the world},
  author={Roy, Deb and Reiter, Ehud},
  journal={Artificial Intelligence},
  volume={167},
  number={1-2},
  pages={1--12},
  year={2005},
  publisher={Elsevier}
}

@article{cho2014learning,
  title={Learning phrase representations using RNN encoder-decoder for statistical machine translation},
  author={Cho, Kyunghyun and Van Merri{\"e}nboer, Bart and Gulcehre, Caglar and Bahdanau, Dzmitry and Bougares, Fethi and Schwenk, Holger and Bengio, Yoshua},
  journal={arXiv preprint arXiv:1406.1078},
  year={2014}
}

@inproceedings{xu2015show,
  title={Show, attend and tell: Neural image caption generation with visual attention},
  author={Xu, Kelvin and Ba, Jimmy and Kiros, Ryan and Cho, Kyunghyun and Courville, Aaron and Salakhudinov, Ruslan and Zemel, Rich and Bengio, Yoshua},
  booktitle={International Conference on Machine Learning},
  pages={2048--2057},
  year={2015}
}

@article{ororbia2017learning,
 author = {Ororbia~II, Alexander G. and Mikolov, Tomas and Reitter, David},
 title = {Learning Simpler Language Models with the Differential State Framework},
 journal = {Neural Computation},
 issue_date = {December 2017},
 volume = {29},
 number = {12},
 year = {2017},
 issn = {0899-7667},
 pages = {3327--3352},
 numpages = {26},
 doi = {10.1162/neco_a_01017},
 acmid = {3186685},
 publisher = {MIT Press},
 address = {Cambridge, MA, USA},
}

@inproceedings{vinyals2015show,
  title={Show and tell: A neural image caption generator},
  author={Vinyals, Oriol and Toshev, Alexander and Bengio, Samy and Erhan, Dumitru},
  booktitle={Computer Vision and Pattern Recognition (CVPR), 2015 IEEE Conference on},
  pages={3156--3164},
  year={2015},
  organization={IEEE}
}

@article{Hochreiter:1997,
 author = {Hochreiter, Sepp and Schmidhuber, J\"{u}rgen},
 title = {Long Short-Term Memory},
 journal = {Neural Computation},
 issue_date = {November 15, 1997},
 volume = {9},
 number = {8},
 month = nov,
 year = {1997},
 issn = {0899-7667},
 pages = {1735--1780},
 numpages = {46},
 url = {http://dx.doi.org/10.1162/neco.1997.9.8.1735},
 doi = {10.1162/neco.1997.9.8.1735},
 acmid = {1246450},
 publisher = {MIT Press},
 address = {Cambridge, MA, USA},
}
